\documentclass{article}
\usepackage{spconf,amsmath,graphicx}
\usepackage{tabularx}
\usepackage{tabularx}
\usepackage{booktabs}

\title{Constrained Convolutional-Recurrent Networks to Improve Speech Quality with Low Impact on Recognition Accuracy}
\twoauthors
  {Rasool Fakoor\sthanks{Work was done as an intern at Microsoft Research Redmond. Correspondence to: Rasool Fakoor ( rasool.fakoor@mavs.uta.edu).}}
	{Dept. of Computer Science and Engineering \\
	Univ. of Texas at Arlington, TX 76019 \\
	\small{\texttt{rasool.fakoor@mavs.uta.edu}\vspace*{-5pt}}
	}
  {\hspace*{-40pt} Xiaodong He, Ivan Tashev, Shuayb Zarar}
	{\hspace*{-40pt} Microsoft Research \\ 
    \hspace*{-40pt} Redmond WA 98052 \\
	\hspace*{-40pt} \small{\texttt{\{xiaohe,ivantash,shuayb\}@microsoft.com}\vspace*{-5pt}}
	}

\begin{document}
%
\maketitle
\begin{abstract}
For a speech-enhancement algorithm, it is highly desirable to simultaneously improve perceptual quality and recognition rate. Thanks to computational costs and model complexities, it is challenging to train a model that effectively optimizes both metrics at the same time. In this paper, we propose a method for speech enhancement that combines local and global contextual structures information through convolutional-recurrent neural networks that improves perceptual quality. At the same time, we introduce a new constraint on the objective function using a language model/decoder that limits the impact on recognition rate. Based on experiments conducted with real user data, we demonstrate that our new context-augmented machine-learning approach for speech enhancement improves PESQ and WER by an additional $24.5\%$ and $51.3\%$, respectively, when compared to the best-performing methods in the literature.

\end{abstract}

\vspace*{-5pt}
\section{Introduction}\label{sec:intro}

Recently, deep learning architectures have led to remarkable progress in problems like speech recognition~\cite{HintonACSP,Dahl2012}, image classification \cite{alexnet2012}, machine translation~\cite{BahdanauCB14, s2sIlya}, image and video caption generation~\cite{FakoorMMKK16, icml2015_xuc15}, speech separation and enhancement \cite{ Samui2017DeepRN, WangLee2017, Kim17IC,fakoorRL} and many others.
Speech enhancement is the process of eliminating noise from an audio signal prior to primarily two higher-level tasks, namely recognition and playback through speaker phones~\cite{Tashev2009}. Because traditional analytical processing methods have a limited capacity to capture complex signal and noise statistics, data-driven approaches are becoming increasingly popular to enhance speech~\cite{XuLeeDU2014,Msamadi2016CSE, Xu2014, AndrewMaas2012}. These learning based approaches typically aim to optimize a particular criterion during training (i.e. the signal mean-squared error (MSE)), while the performance of speech enhancement is usually evaluated from different aspects by multiple metrics (e.g. WER, PESQ) during test and inference time. Therefore, there is a metric discrepancy between training and evaluation, which leads to suboptimal performance.

Jointly training the speech enhancement and recognition systems (i.e. ASR) to simultaneously improve MSE and WER could potentially alleviate this problem. Unfortunately, not only optimizing such a model can be extremely challenging namely due to the complexity of ASR models but also it can be computationally very expensive, raising the need for careful modeling and training. Motivated by these observations, in this paper, we propose a model that not only effectively combines global and local contextual knowledge to enhance speech but also learns how to regularize the speech enhancement and denoising model such that the metric discrepancy could be mitigated. Specifically, to achieve good enhancement performance, our proposed model consists of convolutional layers coupled with recurrent cells. Further, we constrain this model by including a language model/decoder in the optimization objective function. Thus, our network tries to limit the impact on recognition rate, while improving speech quality. To effectively train our model, we also adapt a curriculum-learning-based{~\cite{Bengio2009CL} training paradigm.

In contrast to our approach, existing methods for speech enhancement utilize a single, unconstrained signal-quality criterion such as the MSE for optimization~\cite{XuLeeDU2014,Msamadi2016CSE, Xu2014, AndrewMaas2012}. Thus, although these algorithms improve speech quality, they degrade the recognition rate (measured by the WER metric). In this paper, we aim to overcome this limitation. The following are the specific contributions that we make:
\vspace*{-3pt}
\begin{itemize}
\itemsep-3pt
\item We propose a contextually-aware neural-network architecture for speech enhancement that is constrained with a language-decoder model to limit the impact on WER.
 \item We demonstrate a methodology to train such a network based on curriculum learning for multi-task regression.
 \item Through extensive experimentation and analysis, we show simultaneous improvements of $24.5\%$ and $51.3\%$ in PESQ and WER over existing methods in the literature that only optimize signal quality. 
\end{itemize}
\vspace*{-3pt}
\vspace*{-0.2in}
\section{Proposed Approach}\label{sec:model}
One of the main challenges in speech enhancement when using deep neural networks is to effectively combine the local and global structure of input frames. For example, the model not only needs to learn how to denoise an audio frame based on the signal in that frame but also needs to take into account the temporal structure of the entire sequence of frames over a short span of time. The recurrent neural network (RNN) is a good fit for the speech enhancement problem given its capability to model the temporal structure of speech data. As we will show in section \ref{sec:exp}, although the RNN provides a useful structure for this problem, it is insufficient to achieve good performance. This is because input speech segments are very long, usually containing thousands of frames, making it difficult for the RNN to catch both local and global contextual information for speech enhancement. 

Moreover, state-of-the-art speech enhancement and denoising models are usually trained on a particular criterion, while the performance is evaluated from different aspects by multiple metrics. For example, most of the models for speech enhancement are formulated as a regression problem~\cite{Msamadi2016CSE, Xu2014}
and use MSE as the loss function during training. However, during the evaluation, PESQ, WER, or sentence error rate (SER) are used to assess the performance of the trained model. There is a significant metric discrepancy between training and testing, \textit{e.g.}, a model that is trained to achieve the lowest MSE during training does not necessarily give an improvement in WER or SER at test time.

To address these problems, we first propose a convolution-recurrent neural network (CRNN) that can efficiently model local and global structure of the speech data. Moreover, we also propose a multi-task learning approach that addresses the metric discrepancy problem and leads to a more robust performance on the speech enhancement and denoising task. 

\vspace*{-0.19in}
\subsection{Combining Local and Global Contexts}
One way to capture temporal structure of the data is to use RNNs to model this relationship. However, simple RNNs do not have the adequate capacity to model both the long-term dependencies and local contextual information among different frames~\cite{Pascanu2012, Hochreiter}. However, for good performance, the model needs to capture the local context among neighboring frames as well as the global context. This is important because the denoising networks not only need to use the surrounding frames to denoise the current frame but also higher-level relationships to build a more effective model.

Motivated by these observations, we propose the CRNN, which models long-term dependencies between frames by the recurrent structure in the network and the local context by applying a convolution network over a local context window of neighboring frames. In this model, at every time step, $t$, our model first utilizes eight neighboring frames as the input to a three-layer convolution network that models the local structure of the input frame ($f_t$). The output of this network will be an input to an LSTM~\cite{Hochreiter} unit at time $t$. The recurrent unit uses the current noise frame as well as previous hidden states to reconstruct an enhanced frame. To be specific, our network at time step $t$ utilizes eight neighboring frames and $h_{t-1}$ to reconstruct a single denoised frame $\hat{g_t}$. Our proposed model is shown in Figure\ref{fig:icms}.

\begin{figure}[tb]
\centering
\centerline{\includegraphics[width=0.26\textwidth]{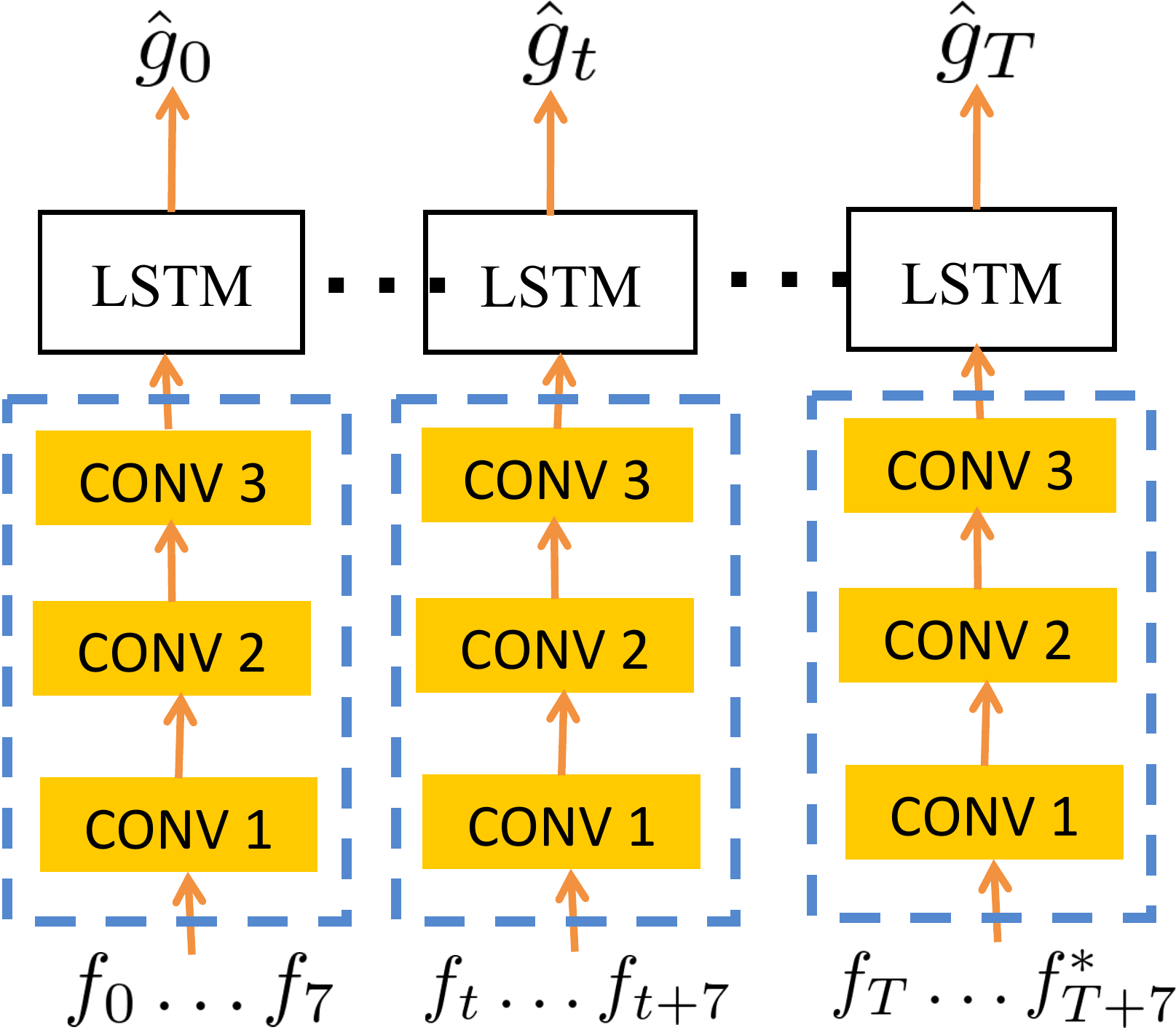}}
\caption{Proposed architecture to combine local and global context of input frames.}
\label{fig:icms}
\end{figure}

The objective function for this model minimizes the error between the enhanced frame (or denoised frame for short) and clean frame which can formally be defined as follow:
\begin{equation}\label{eq:RE}
L_{re}(F;\mathbf{\omega}) = min_{\omega}\sum_t \parallel \hat{g_t} - g_t \parallel_2^2  
\end{equation}
where $\omega$ is a network parameter, $\hat{g_t}$ is a denoised frame, and $g_t$ is the clean frame that we use to train the parameters of this model. As we show in the experimental results section, this model outperforms other networks that either only model the local structure or the global structure of the data. 
\vspace*{-0.14in}
\subsection{Multi-task Learning}

The metric discrepancy is also the main challenge that most of the speech enhancement and denoising models face ~\cite{Msamadi2016CSE, Xu2014, AndrewMaas2012}. For example, while most of the models use MSE as the training metric, the performance is evaluated on other criteria such as PESQ, SER, and WER.
One possible solution is to train the models to directly optimize PESQ or WER. Although this is plausible, there are a couple of problems. First, these metrics are very expensive to calculate and often it is not practical to use them directly during training. Moreover, these metrics (i.e PESQ) are usually discrete and non-differentiable, making the optimization (Eq. \ref{eq:RE}) very difficult. The REINFORCE algorithm \cite{Williams} can be used in certain situation, but gradient estimation using REINFORCE would be non-trivial in this setup as it deals with a continuous number for sampling and gradient calculation (i.e. at each time step, network outputs a denoised frame which is a continuous vector in the speech space, making the policy-gradient estimating a hard problem~\cite{icml2014c1_silver14}).

Motivated by these problems, we propose a multi-task learning framework during training that uses a language model (i.e. language decoder) to regularize the training and improve the performance of the denoising network with respect to PESQ, WER, or SER. That is, we first run the model to perform denoising, and once the denoising is done, i.e., reaches the last frame of the input speech segment, our approach uses the last hidden unit representation of the CRNN as the input to a RNN based language model, in which the language model is trained to generate the text transcript of the input speech segment. This is like imposing another task of sequence-to-sequence multimodal translation, which encodes the sequence of the denoised speech signal into one vector representation and then translates it into the sequence of words in the transcription. In order to build this language model, we add the following loss function:
\begin{equation}\label{eq:NLL}
L_{lm}(S, H_t; \mathbf{\theta}) = -\sum_{j^\prime}^T \sum_{i}^{|V|} s_i^{j^\prime}\log (\hat{s}_i^{j^\prime}) + \lambda\parallel\Theta \parallel_2^2 
\end{equation} 
where $H_T$ is the last hidden unit of the denoising model, $S$ is a transcript for a given file, and $\theta$  is the parameter of this network. The Eq. \ref{eq:NLL} is cross-entropy loss function that tries to minimize the word prediction error. Combing this loss function with Eq. \ref{eq:RE} helps the network to constraint and regularize the denoising network such that it will have better performance regarding WER, SER, and PESQ during test time:
\begin{equation}\label{eq:RE_NLL}
L(S, F, H_t;\mathbf{\Theta}) = L_{re} + \lambda_{1}L_{lm} + \lambda_{2}\parallel\Theta \parallel_2^2 
\end{equation}
This architecture is shown in Figure~\ref{fig:lm} in which the language model is shown in dotted box $(b)$ and denoising model is shown in dotted box $(a)$. It is worth noting that the intuition behind this model is that the original denoising model does unconstrained optimization\footnote{ By unconstrained optimization, we are referring to the fact that we did not explicitly impose any constraints on Eq. \ref{eq:RE}, i.e. bound the model outputs} as the result, the denoising model only minimizes the MSE as much as it can without considering PESQ, WER, etc. This causes the model to sometimes overfit on the MSE metric and shows worst performance on other metrics. However, by adding the language model, the model is not only focused on minimizing the MSE, but also tries to denoise in a way such that the denoised speech signal can lead to better word prediction decoded by the language model. Therefore, adding the language modeling task effectively regularize the training of the denoising model and will lead to more robust performance as reflected in the improvements in terms of WER and PESQ too. As the results show, this approach is very effective, outperforms other methods significantly on a range of evaluation metrics.

\begin{figure}[tb]
\centering
\centerline{\includegraphics[width=0.5\textwidth]{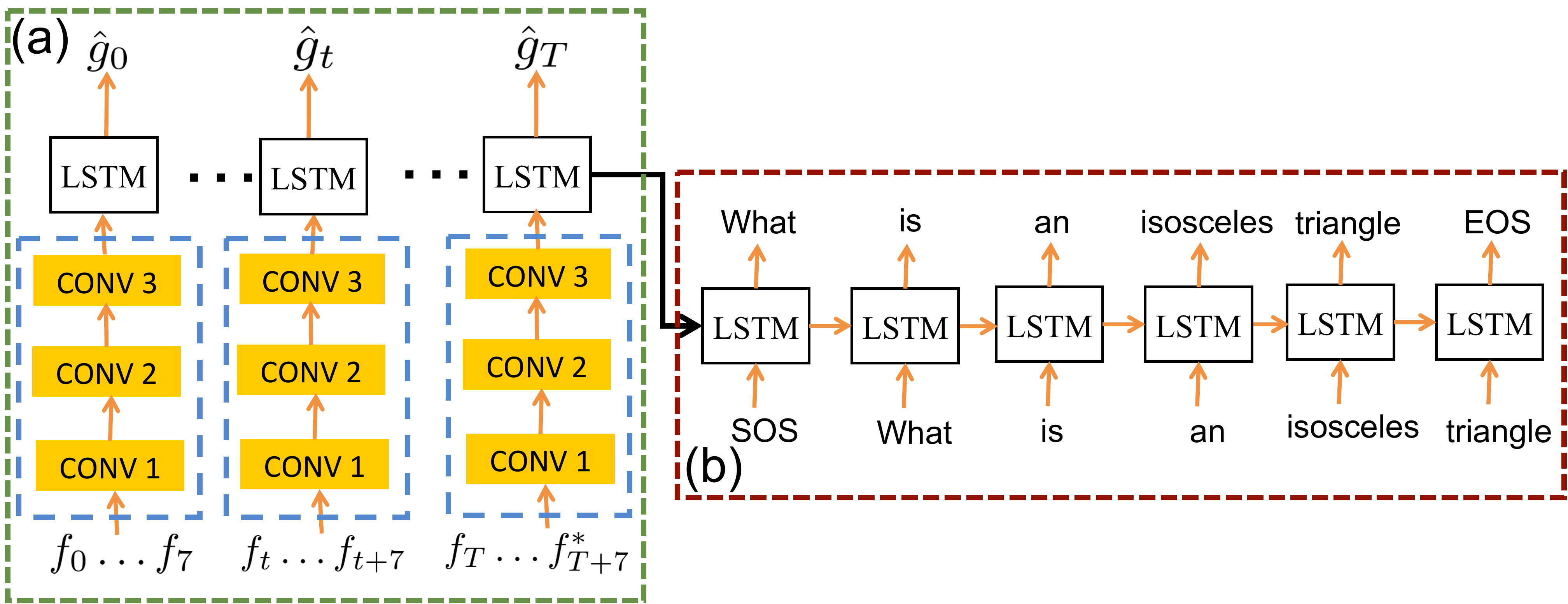}}
\caption{Our multi-tasks based learning architecture.}
\label{fig:lm}
\end{figure}

\vspace*{-0.14in}
\subsection{Curriculum Learning}\label{sec:cl}
The language model and the denoising model operate very differently, while the language model catches the dependency at the word level, the denoising model works at the lower speech frame level. If we train them together from scratch, the model has hard time to converge. Specifically, the denoising model needs hundreds of epochs to converge to a stable model given the level of difficulty and complexity in this denoising problem. On the other hand, since there is one transcript per file and usually it is short (i.e around 60 words per files),  the language model only needs a few epochs to converge.  
To deal with this problem, we design a curriculum learning paradigm~\cite{Bengio2009CL} to train this model. We first train the denoising model for few hundred epochs until it stops improving, i.e. only train it with Eq.\ref{eq:RE}. At this stage, we introduce Eq. \ref{eq:RE_NLL} to the model as the new objective functions. Note that our proposed curriculum learning is different from traditional one~\cite{Bengio2009CL} such that in the traditional curriculum learning, it first starts with the simpler problem then goes to the harder problem. However, in our proposed approach, we first start with the core task, then we combine it with another task to further regularize the training. As shown in the next section, the proposed method is very effective for the challenging denoising task.
\vspace*{-18pt}
\section{Experiments}\label{sec:exp}

\begin{table*}[t]
\small
\centering
    \begin{tabularx}{.91\textwidth}{llllllllll}
      \toprule 
      Method & {\sc SNR(dB)} & {\sc LSD} &{\sc MSE} & {\sc SIR} & {\sc SDR} & {\sc SAR} & {\sc WER} & {\sc SER} & {\sc PESQ}\\ 
            \midrule
      \textbf{Noisy data}      & $15.18$ & $23.07$ & $0.04399$ & $39.1$ & $-0.67$ & $-0.66$ & $15.4$ & $25.07$ & $2.26$ \\
\textbf{Existing model~\cite{AndrewMaas2012}} & $41.08$ & $17.49$ & $0.03533$ & $8.58$ & $-0.58$ & $1.68$ & $44.93$ & $66.60$ & $2.19$ \\
\textbf{Existing model~\cite{Msamadi2016CSE}} & $40.70$ & $20.09$ & $0.03485$ & $7.47$ & $-1.28$ & $1.32$ & $54.92$ & $75.87$ & $2.17$ \\

\textbf{Existing model~\cite{Msamadi2016CSE}} & $44.51$ & $19.89$ & $\mathbf{0.03436}$ & $7.84$ & $-1.04$ & $1.48$ & $55.38$ & $74.93$ & $2.20$ \\
\textbf{Existing model~\cite{Xu2014}}     & $27.03$ & $20.84$ & $0.03711$ & $5.82$ & $-2.36$ & $0.80$ & $60.72$ & $79.87$ & $1.93$ \\
\midrule
\textbf{Our model (NOC)} & $40.23$ & $17.78$ & $0.03544$ & $6.84$ & $-1.50$ & $1.23$ & $45.19$ & $66.40$ & $2.23$ \\
\textbf{Our model (CRNN)} & $41.26$ & $15.90$ & $0.03480$ & $9.52$ & $0.19$ & $2.12$ & $22.73$ & $38.93$ & $2.70$ \\
\midrule
\textbf{Our model (CRNN + LM)} & $44.22$ & $15.94$ & $0.03462$ & $9.48$ & $0.18$ & $2.14$ & $23.79$ & $40.13$ & $2.69$ \\ 
\textbf{Our model (CRNN + LM +CL)} & $40.38$ & $16.14$ & $0.03457$ & $\mathbf{9.76}$ & $\mathbf{0.36}$ & $\mathbf{2.23}$ & $\mathbf{21.90}$ & $\mathbf{37.27}$ & $\mathbf{2.74}$ \\
\midrule
\textbf{Clean data}      & $57.31$ & $1.01$ & $0.0$ & $79.02$ & $57.05$ & $58.36$ & $2.19$ & $7.4$ & $4.48$ \\
    \end{tabularx} 
        \caption{Performance comparison for speech enhancement tasks. \label{tab:main_res}}
\end{table*}
\vspace*{-0.2in}
\subsection*{Dataset}
We evaluate the performance of our methodology with single-channel recordings based on real user queries to the Microsoft Windows Cortana Voice Assistant. We split studio-level clean recordings into training, validation and test sets comprising $7500$, $1500$ and $1500$ queries, respectively. Further, we mixed these clean recordings with noise data (collected from $25$ different real-world environments), while accounting for distortions due to room characteristics and distances from the microphone. Thus, we convolved the resulting noisy recordings with specific room-impulse responses and scaled them to achieve a wide input SNR range of $0$-$30$ dB. Each (clean and noisy) query had more than $4500$ audio frames of spectral amplitudes, each lasting $16$ ms. We applied a smoothing function based on a Hann window to the frames allowing accurate reconstruction with a $50\%$ overlap. These audio frames in the spectral domain formed the features for our algorithm. Since we utilized a $512$-point short-time Fourier Transform (STFT), each feature vector was a positive real number of a dimensionality of $256$.
\vspace*{-0.2in}
\subsection*{Hyperparameter Optimization}
We use random search \cite{Bergstra2012} on the validation set
to select hyperparameters for this dataset. A stack of two LSTMs are used in our best model (\textbf{CRNN + LM} and \textbf{CRNN}). These models both have $1072$ hidden units. The weight decays are $2.8951e^{-5}$ and $3.6998e^{-5}$. In addition, the \textbf{LM} uses $857$ words in its vocabulary and all transcript are capped to have $60$ words maximum. In order to optimize our network parameters, we use Adam~\cite{KingmaB14} with learning rates of $6.4710e^{-5}$ and set $\beta 1$, $\beta2$ to $0.8$ and  $0.999$, respectively. The convolution layers in our models (yellow boxes in Fig. \ref{fig:icms} and \ref{fig:lm}) have the following specifications: 
1) Conv 1 has $16$ filters with the kernel size of $(7, 5)$, stride size of $(3, 1)$, 2) Conv 2 has $32$ filters with the kernel size of $(5, 3)$, stride size of $(3, 1)$, and 3) Conv 3 has $64$ filters with the kernel size of $(5, 1)$, stride size of $(3, 1)$. In addition, all convolution layers use $(2, 1)$ dilation~\cite{YuK15} as well.
\vspace*{-0.18in}
\subsection*{Performance Comparison}
We carry out an extensive evaluation to evaluate the proposed models. In the evaluation, we compare the proposed model with state-of-the-art baselines for the speech enhancement and denoising task. We summarize the results of these experiments in Table \ref{tab:main_res}. We compare our models to recent deep neural network based approaches which are strong baselines, including ~\cite{Xu2014}, \cite{Msamadi2016CSE}, and \cite{AndrewMaas2012}.  
We first build a model (\textbf{NOC}) that does not consider the global context of the data (i.e. no RNN) and only considers local context. Then we extend these models to our denoising model \textbf{CRNN}.
As the results show, our proposed model \textbf{CRNN} outperforms the baselines on the key metrics of PESQ, WER, SER, and others. 

In addition, Table \ref{tab:main_res} shows that our proposed multi-task model (\textbf{CRNN + LM}) outperforms other models and furthers improve PESQ, WER, and SER. In addition, we show in Figure \ref{fig:pseq_incr} the improvement in PESQ scores by using our model.
\begin{figure}[ht]
 \centering
  \centerline{\includegraphics[width=0.8\linewidth]{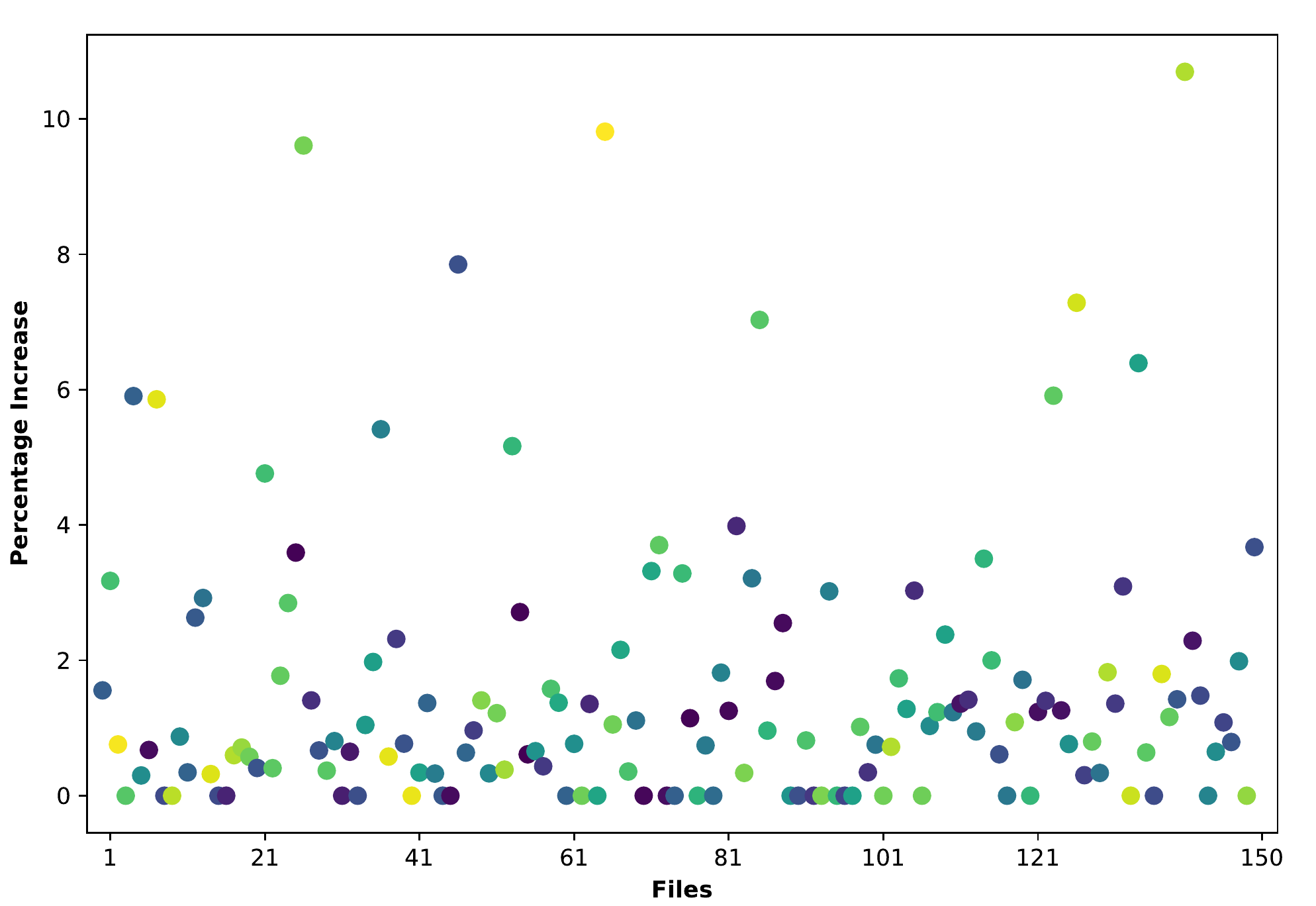}}
    \caption{The positive percentage increase for PESQ score in the test files. }
\label{fig:pseq_incr}
\end{figure}

\vspace*{-0.379in}
\subsection*{Curriculum learning}
We also studied the impact of the proposed curriculum learning procedure. Given the large gap between denoising model which operate at the lower speech frame level and the language model which operate at the higher word level, it is important to use the proposed curriculum learning based training method (section ~\ref{sec:cl}) to train our model. As results in Table show \ref{tab:res_cl}, when we train both denoising network and language model together from the beginning, the performance is quite bad, compared to the performance using the proposed curriculum learning.  

\begin{table}[ht]
\small
\centering
    \begin{tabularx}{0.95\columnwidth}{llll}
      \toprule 
      Method & {\sc WER} & {\sc SER} & {\sc PESQ}\\ 
            \midrule
\textbf{Our model (CRNN)} & $22.73$ & $38.93$ & $2.70$ \\
\textbf{Our model (CRNN + LM)} & $23.79$ & $40.13$ & $2.69$ \\
\textbf{Our model (CRNN + LM + CL)} & $\mathbf{21.90}$ & $\mathbf{37.27}$ & $\mathbf{2.74}$ \\
\end{tabularx} 
        \caption{Results on effects of Curriculum learning (CL).\label{tab:res_cl}}
\end{table}

\vspace*{-0.22in}
\section{Conclusion}\label{sec:con}
In this paper, we propose a model that combines both local and global contextual information for speech enhancement. We show that our approach leads to better enhancement performance compared to existing baselines. Furthermore, we propose multi-task learning with curriculum learning, which regularizes the training process of the speech-enhancement model through a language model/decoder. Thus, we limit the impact of speech enhancement on recognition accuracy.

\newpage
\bibliographystyle{IEEEbib}
\bibliography{refs}

\end{document}